\pdfoutput=1

\documentclass[11pt]{article}

\usepackage[final]{acl}

\usepackage{times}
\usepackage{latexsym}

\usepackage[T1]{fontenc}

\usepackage[utf8]{inputenc}

\usepackage{microtype}

\usepackage{inconsolata}

\usepackage{booktabs}
\usepackage{multirow}
\usepackage{graphicx}
\usepackage{makecell}
\usepackage{xcolor}
\usepackage{amsmath}

\title{Enhancing Temporal Sensitivity and Reasoning for Time-Sensitive Question Answering}

\author{
 \bfseries
 Wanqi Yang$^{1 \ast}$,
 \
 Yanda Li$^{1}$\thanks{Equal contributions},
 \
 Meng Fang$^{2}$,
 \
 Ling Chen$^1$
 \\
 \normalsize 
 $ ^1$ University of Technology Sydney
 \
 $ ^2 $ University of Liverpool
 \\
 {$^1$\normalsize \tt   wanqi.yang-1@student.uts.edu.au, 
 $^{1}$Yanda.Li@student.uts.edu.au}\\
 $^2$\normalsize \tt  Meng.Fang@liverpool.ac.uk,
 $^1$\normalsize \tt ling.chen@uts.edu.au
 }

\begin{document}
\maketitle
\begin{abstract}
Time-Sensitive Question Answering (TSQA) demands the effective utilization of specific temporal contexts, encompassing multiple time-evolving facts, to address time-sensitive questions. This necessitates not only the parsing of temporal information within questions but also the identification and understanding of time-evolving facts to generate accurate answers. 
However, current large language models still have limited sensitivity to temporal information and their inadequate temporal reasoning capabilities.
In this paper, we propose a novel framework that enhances temporal awareness and reasoning through Temporal Information-Aware Embedding and Granular Contrastive Reinforcement Learning. 
Experimental results on four TSQA datasets demonstrate that our framework significantly outperforms existing LLMs in TSQA tasks, marking a step forward in bridging the performance gap between machine and human temporal understanding and reasoning.
\end{abstract}

\section{Introduction}

Time-Sensitive Question Answering (TSQA) involves parsing and responding to questions that depend on specific time points or periods. For instance, Obama's roles in 2006 and 2016 were distinctly different. Unlike conventional Question Answering (QA) tasks~\cite{rajpurkar2016squad,joshi2017triviaqa,dunn2017searchqa,ye2023multi,zhao2023verify}, TSQA requires language models to discern and understand temporal information within questions.  Moreover, it necessitates the capacity to locate relevant facts using temporal information, leveraging time-related knowledge to provide accurate and relevant answers. In the domain of TSQA, models are required to predict answers through generation or extraction methods, given a set of questions and contexts embedded with temporal information. These questions are characterized by explicit or implicit temporal expressions, while the context invariably contains multiple facts evolving over time. Recently, several TSQA datasets have been introduced, including notable work~\cite{chen2021dataset} that provides 40k time-sensitive questions with corresponding free-text contexts derived unprocessed from WikiData~\cite{vrandevcic2014wikidata}. Building upon this foundation, ~\cite{tan2023towards} not only furnish free-text contexts from WikiData without any modification but also offer a version of the dataset where the context has been restructured. These contributions have facilitated evaluations of existing large language models (LLMs) such as FiD~\cite{izacard2020leveraging}, BigBird~\cite{zaheer2020big}, and T5~\cite{raffel2020exploring} on these datasets, underscoring the challenge TSQA poses, as the performance of LLMs remains significantly inferior to human levels. However, an effective methodology to significantly improve LLMs performance in TSQA is still lacking. 

It is challenging to improve model's sensitivity to temporal information and the capacity for temporal reasoning.
Firstly, LLMs significantly lack attention to and comprehension of temporal information in both questions and their contexts~\cite{ning2020torque,shang2021open}.
 It is essential for models to prioritize temporal cues in questions to accuately locate corresponding times and facts within contexts, 
 especially for lengthy and complex free-text contexts, which encompass multiple irrelevant information and various chronological connections. Focusing on temporal information allows a model to circumvent the distraction of irrelevant data and to swiftly locate relevant facts through the chronological connections among different paragraphs. 
Secondly, LLMs~\cite{han2020econet,dhingra2022time} exhibit notably weak temporal reasoning abilities. This specifically refers to the models' susceptibility to interference from answers that pertain to the same entity or relationship but span different time periods within the context, as well as to disturbances from other events occurring within the same time period.

In this paper, we introduce a framework to address these challenges. Initially, we implement Temporal Information-Aware Embedding to enhance the model's sensitivity towards temporal information by augmenting its focus on temporal data and adjacent specifics. 
In our time-aware module, we use SpaCy and identify temporal information within questions and contexts, establishing a question temporal matrix and a context temporal matrix. 
Furthermore, we advance the model's temporal reasoning capabilities through the Granular Contrastive Reinforcement Learning. We introduce remote negative answers — answers within contexts corresponding to different time periods for the same entities and relationships, and proximal negative answers — answers related to other events within the same time period. Additionally, we propose a more rational reward function to aid the model in reinforcement learning. Conclusively, we present comparative results between our framework and existing LLMs, validating our framework's effectiveness through its application across four TSQA datasets with varying difficulty.

The main contributions of this work are summarised as:
 \begin{itemize}
 \vspace{-1mm}
    \item We introduce a new TSQA framework that explicitly processes time-aware information to significantly enhance the performance of LLMs in TSQA tasks.
\vspace{-1mm}    
    \item We propose a time-aware module for constructing Temporal Information-Aware Embedding to enhancing the model's sensitivity to temporal information.
\vspace{-1mm}    
    \item We propose Granular Contrastive Reinforcement Learning to assist the model in improving its temporal reasoning capabilities.   
\vspace{-1mm}    
    \item We present experimental results that demonstrate our framework significantly outperforms existing LLMs across various TSQA datasets.
\end{itemize}
\begin{figure*}[h]
\centering
  \includegraphics[width=0.95\textwidth]{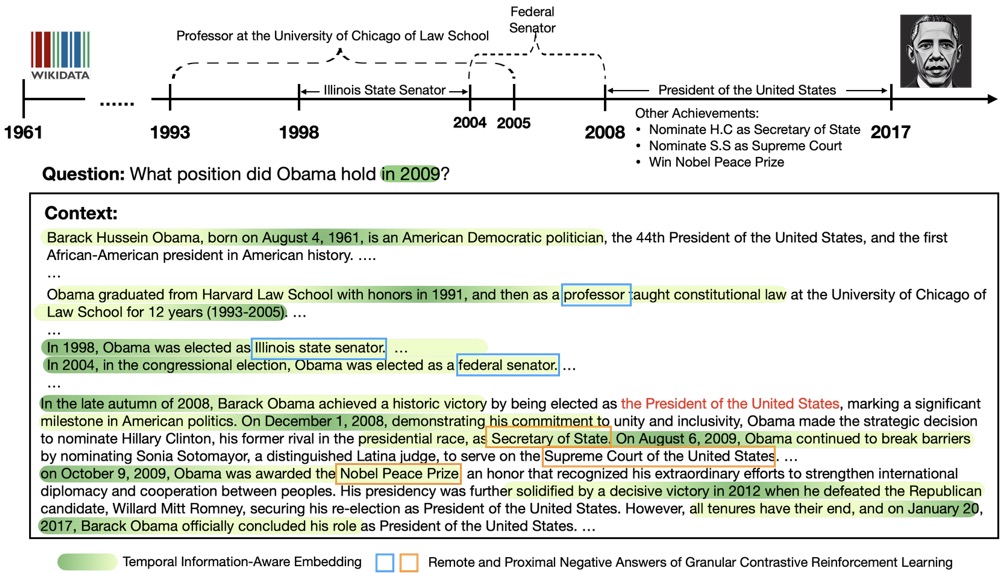}
  \caption{An overview of the TSQA task with our framework. }
\vspace{-4mm}
    \label{method}
 \end{figure*}

\section{Problem Definition}

Time Sensitive Question Answering aims to generate answers $\tilde{A}$ based on a given free-text question $Q$ and its corresponding free-text context $C$. The questions feature explicit or implicit temporal expressions, while the context invariably includes multiple time-evolving facts ($s$, $r$, $o$, [$t_s$, $t_e$]) where $r$ denotes the relationship that exists between the subject entity $s$ and the object entity $o$ throughout the specified time interval [$t_s$, $t_e$]. The answer is either an entity or non-existent.
The characteristics of TSQA include: 1) The structure ($s$, $r$, $o_i$, [$t_{s_i}$, $t_{e_i}$]) is invariably extractable from the question and the answer. 2) Modifying the temporal information within the question will result in a change to the answer.

The TSQA primarily evaluates the model's sensitivity to temporal information and its ability of temporal reasoning. \textbf{Temporal information sensitivity} refers to the model's capability to pay closer attention to the temporal information presented in both the given question and the corresponding context, recognizing the temporal information as a crucial factor. \textbf{Temporal reasoning ability} denotes the model's understanding of basic temporal relationships (for example, The year 2015 falls within the time interval from 2010 to 2020, the year 2015 is five years later than the year 2010, the period of World War II is within the time interval from 1939 to 1945, etc.) and its capacity to deduce the correct answer based on these temporal relationships.

\section{Method}
To enhance the sensitivity to temporal information and the capability for temporal reasoning within models, we propose a training framework. Our framework addresses the TSQA task by leveraging a pre-trained language model to interpret free-text temporal questions and context. As illustrated in Fig.~\ref{method}, the framework incorporates two primary methodologies: Temporal Information-Aware Embedding and Granular Contrastive Reinforcement Learning.

Temporal Information-Aware Embedding is dedicated to enhancing the model's sensitivity to temporal information by increasing its attention to temporal data and adjacent temporal details. Granular Contrastive Reinforcement Learning offers remote and proximal negative answers based on varying temporal distances and employs a more rational reward function to assist the model in improving its temporal reasoning capabilities.

\subsection{Temporal Information-Aware Embedding}

\begin{figure*}[h]
\centering
  \includegraphics[width=0.8\textwidth]{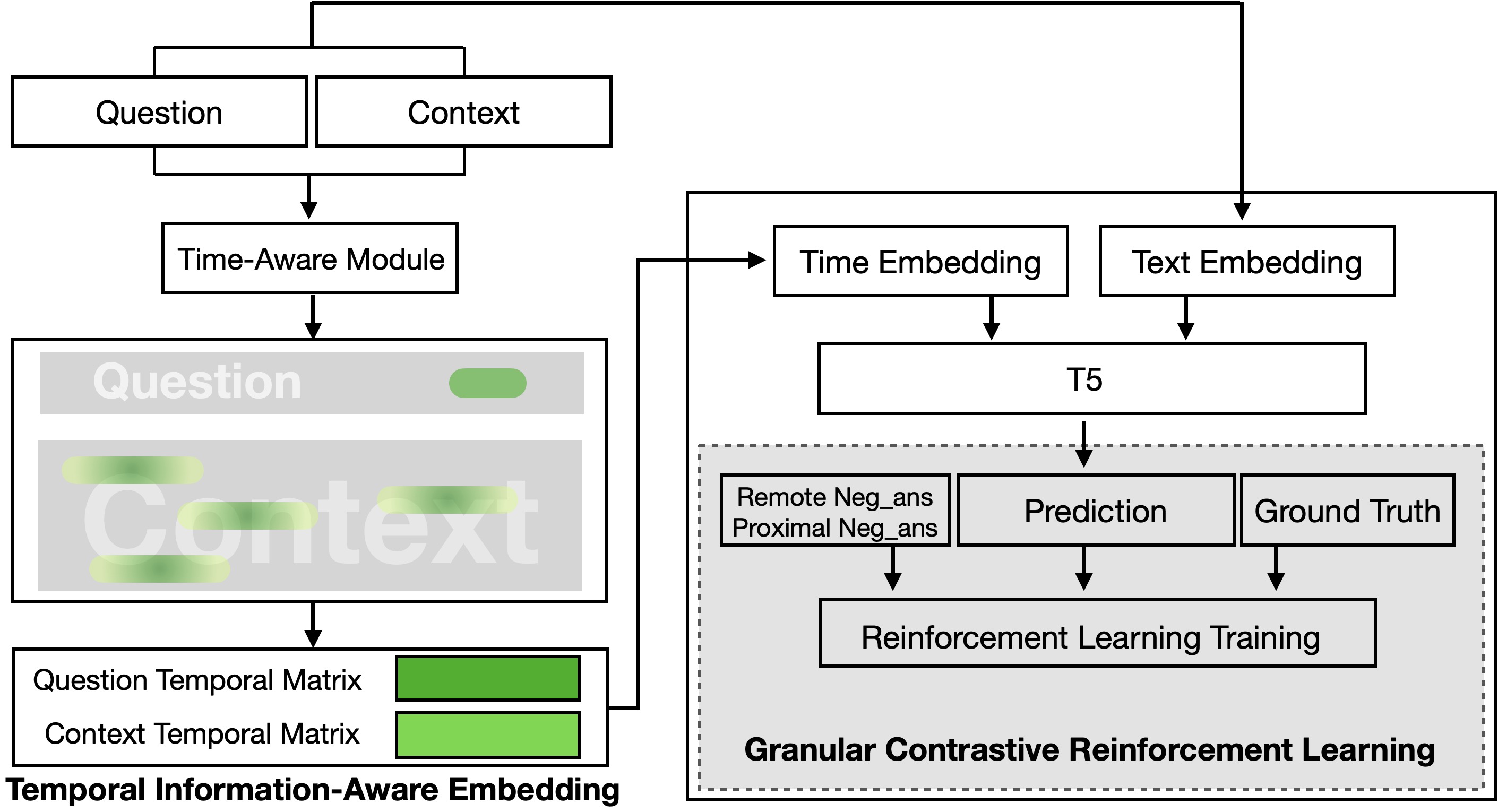}
\vspace{-1mm}
  \caption{The architecture of our framework (Left: Temporal Information-Aware Embedding;Right: Granular Contrastive Reinforcement Learning).}
\vspace{-5mm}
    \label{network}
 \end{figure*}

In reading comprehension tasks with temporal questions, humans typically begin by identifying temporal cues within the question and subsequently locating corresponding cues within the context. These cues often guide them to accurate answers situated near these temporal markers. Inspired by this intuitive human approach, we propose the Temporal Information-Aware Embedding. This method is specifically designed to enhance the sensitivity of LLMs to temporal details and their associated contextual information. The implementation details of this technique are illustrated on the left side of Figure~\ref{network}, and include the following steps:


1) Firstly, we construct a question temporal matrix $A_q$ and a context temporal matrix $A_c$, both initially filled with zeros as:
\vspace{-2mm}
\begin{equation}
    A_q=\begin{bmatrix}
    a_{q_1} & \dots & a_{q_n}\\
    \end{bmatrix}_{1 \times n}
    =
    \begin{bmatrix}
    0 & \dots & 0\\
    \end{bmatrix}_{1 \times n}
\end{equation}
\vspace{-4mm}
\begin{equation}
    A_c=\begin{bmatrix}
    a_{c_1} & \dots & a_{c_m}\\
    \end{bmatrix}_{1 \times m}
    =
    \begin{bmatrix}
    0 & \dots & 0\\
    \end{bmatrix}_{1 \times m}
\end{equation}

In our time-aware module, we use SpaCy to locate temporal expressions from question $a_{q_t}$ and context $a_{c_t}$, marking these detected positions in $A_q$ and $A_c$ with ones:
\vspace{-1mm}
\begin{equation}
    A_q =\begin{bmatrix}
    a_{q_1} \dots &\boldsymbol{a_{q_t}}& a_{q_n} 
    \end{bmatrix}_{1 \times n} \\
    =
    \begin{bmatrix}
    0 & \dots &\textbf{1}& 0\\
    \end{bmatrix}_{1 \times n}
\end{equation}
\vspace{-3mm}
\begin{equation}
\begin{split}
    A_c &=\begin{bmatrix}
    a_{c_1} & \boldsymbol{a_{c_t}} &a_{c_i}\dots &\boldsymbol{a_{c_t}}&a_{c_j} &\dots a_{c_n} 
    \end{bmatrix}_{1 \times m} \\
    &=
    \begin{bmatrix}
    0 &\textbf{1} &0 & \dots &\textbf{1}&0\dots & 0\\
    \end{bmatrix}_{1 \times m}
\end{split}
\end{equation}
2) Employing a sliding window $W$ with window size $L$ centered on the temporal information, we use $f(W)$ function to mark the positions within the window region as ones  to further highlight regions of temporal relevance. The definition of $W$ for an element $a_i$ is given by:
\vspace{-2mm}
\begin{equation}
    W_{a_i}=\{a_{i-L},\dots, a_i, \dots, a_{i+L}\}
\end{equation}
\vspace{-4mm}

\noindent Consequently, the function \( f(W_{ai}) \) is defined as:

\vspace{-4mm}
\begin{equation}
\begin{split}
    f(W_{a_i}) &=f\{a_{i-L},\dots, a_i, \dots, a_{i+L}\} \\
    &= 1, \quad \forall a_j = 1, \quad j \in [i-L, i+L]
\end{split}
\end{equation}

3) The question temporal matrix and context temporal matrix are concatenated, followed by an embedding layer $\mathbf{W_{time}}$, resulting in the temporal information-aware embedding $\mathbf{e_{time}}$.

4) The question and context, serving as model inputs, are processed through another embedding layer $\mathbf{W_{text}}$ to obtain text embedding $\mathbf{e_{text}}$. This text embedding is then combined with its temporal information-aware embedding, aiming to enhance the model's focus on temporal information and its adjacent information.

This approach systematically embeds temporal awareness into the model, enhancing its ability to understand and prioritize temporal information, thereby improving model performance on temporal questions.

\subsection{Granular Contrastive Reinforcement Learning}
\label{sec:3.3}
As illustrated in right part of Fig~\ref{network}, we propose Granular Contrastive Reinforcement Learning, composed of  Negative Answers with Different Granularity and Contrastive Reinforcement Learning. This methodology enhances the model's temporal reasoning capabilities by assisting in the filtration of negative answers with varying degrees of granularity.

\paragraph{Negative Answers with Different Granularity.}
Accurate temporal reasoning depends on discerning the relevance between potential incorrect answers and the fact. We categorize negative answers into two types based on their temporal and contextual relationship to the fact: Remote Negative Answers and Proximal Negative Answers.

Remote Negative Answers are those that share the same subject and relation with the fact but belong to distinctly different time periods. These answers, although factually related to the question, are temporally distant from the ground truth and thus less likely to be confused with it. For example, considering the question ``What position did Obama hold in 2009?'' with the subject-relational context of Obama's career positions spanning 2008 to 2017, Remote Negative Answers would include positions such as \{Professor at the University of Chicago Law School (1993-2005), Illinois State Senator (1998-2004), Federal Senator (2004-2008)\}. These roles, while relevant, occurred outside the specified time period of the question.

Proximal Negative Answers, in contrast, are selected from different subjects or relations but fall within the same time period as the ground truth. These answers are temporally proximal but contextually distinct, providing a different type of challenge in distinguishing correct answers. Continuing with the Obama example, Proximal Negative Answers might be \{Secretary of State, Supreme Court, Nobel Peace Prize\}, all significant roles or honors from the same period (2008-2017) but unrelated to Obama's positions.

By incorporating negative answers that vary in temporal proximity and contextual relevance, our model is better equipped to understand and discriminate the details of temporal relations in text, enhancing its predictive accuracy in temporally-focused tasks. More examples are shown in Appendix~\ref{sec:appendix2}.




\paragraph{Contrastive Reinforcement Learning.}

we propose a novel contrastive reinforcement learning framework, building on previous methods like the EM algorithm~\cite{tan2023towards}, which occasionally fails to recognize semantically equivalent responses due to its rigid scoring mechanics. For instance, when the model's predicted answer is ``the capital of France'' and the ground truth is ``Paris'', the EM algorithm assigns a score of 0. Nonetheless, it is widely acknowledged that ``the capital of France'' and ``Paris'' should be considered equivalent answers. 

Reinforcement learning focuses on maximizing the likelihood of the correct answer while penalizing temporally inaccurate predictions. Unlike traditional approaches that may rely on direct string comparison, our method employs a vectorial representation of answers to administer the triplet loss as a reward function as below. Our model dynamically evaluates these vectors against a ground truth $GT$, the model's prediction $P$, and a set of negative answers $N$. We have 
\vspace{-2mm}
\begin{equation}
    T = max\{d(GT, P)-d(P, N)+margin, 0 \}
\end{equation}
where $d(x,y) = \parallel x-y \parallel_2$.
The resultant score $T$ is then normalized and scaled, transforming it into a reward function $R$ that is contingent upon the distinctiveness of the model's prediction from the negative samples and its alignment with the ground truth. The reward function is defined as:
\begin{equation}
    R =  \alpha \cdot \left( \frac{2}{1+e^{T}+\delta} \right) - \beta
\end{equation}
where $\alpha$ = 4, $\beta$ = 2 and $\delta$ = $1e^{-6}$.

This approach not only refines the model's ability to make temporally precise judgments but also enhances its overall learning efficiency by dynamically adjusting to the complexities of temporal reasoning. Thereby significantly boosting model's learning efficiency and effectiveness. The model is optimized using Proximal Policy Optimization~\cite{schulman2017proximal}, whose details are shown in Appendix~\ref{sec:appendix1}.

\section{Experiments}

\subsection{Experimental Setup}

\paragraph{Dataset}
\begin{itemize}
\vspace{-1mm}
    \item Time-Event QA (L2) is derived from the TempReason ~\cite{tan2023towards} dataset. The distinguishing features of L2 include: 1) The presence of a specific time point in each question, denoted by ``in YEAR''; 2) The time point mentioned in the question \textbf{may not} find a direct match within the context; 3) Context is provided from Wikidata (OBQA); 4) A structured context is offered, where the content is strictly organized according to the ($s$, $r$, $o$, [$t_s$, $t_e$]) template (ReasonQA).
\vspace{-1mm}
    \item Event-Event QA (L3) also originates from TempReason. L3 is characterized by: 1) \textbf{An absence} of explicit temporal information in the questions, with all time information being replaced by an event (for example, the period from 1939 to 1945 is referred to as the World War II era); 2) Complex expressions of time such as ``before'', ``after'', ``during'', and ``simultaneous'' are used in questions; 3) Context is provided from Wikidata (OBQA); 4) A structured context is offered, where the content is strictly organized according to the ($s$, $r$, $o$, [$t_s$, $t_e$]) template (ReasonQA).
\vspace{-1mm}
    \item TimeQA Easy is based on the TimeQA dataset ~\cite{chen2021dataset}. Features of the Easy Version include: 1) Each question contains a time expression either in the format ``in YEAR'' or ``from YEAR to YEAR''; 2) The time mentioned \textbf{can be} directly matched within the context; 3) Context from Wikidata is provided (OBQA).
\vspace{-1mm}
    \item TimeQA Hard stems from the TimeQA dataset. The Hard Version is characterized by: 1) The presence of time information in questions expressed in complex terms such as ``before'', ``after'', ``first'', ``last''; 2) The time mentioned in the question \textbf{cannot} find a direct match within the context; 3) Context from Wikidata is provided (OBQA).
\end{itemize}

These four types of TSQA present varying levels of difficulty in terms of temporal information expression and temporal reasoning. We investigated the sensitivity to temporal information and the temporal reasoning capabilities of the model using these TSQA datasets with differing difficulties, to demonstrate the outstanding performance of our framework. Table~\ref{datasets} enumerates the number of questions for the four types of TSQA across the training set, validation set, and test set.

\begin{table}[h]
\centering
\resizebox{\linewidth}{!}{
\begin{tabular}{l|c|c|c|c}
\toprule
{Dataset} & {Question Type} & {Train} & {Dev} & {Test} \\ 
\midrule
\multirow{2}*{TempReason} &{L2: Time-Event} & 16017 & 5521 & 5397 \\
\cline{2-5}
\multirow{2}*{} & {L3: Event-Event} & 13014 & 4437 & 4426 \\
\cline{1-5}
\multirow{2}*{TimeQA} &{Easy}& 14308 & 3021 & 2997 \\
\cline{2-5}
\multirow{2}*{} & {Hard} & 14681 & 3087 & 3078\\
\bottomrule
\end{tabular} 
}

\caption{\label{datasets}
The dataset statistics for different TSQA dataset, TempReason and TimeQA. 
}
\vspace{-3mm}
\end{table}

\begin{table*}[t]

\centering
\resizebox{\linewidth}{!}{
\begin{tabular}{l|c|c|c|c|c|c|c|c|c|c}
\toprule
\multirow{2}*{Dataset} & \multirow{2}*{Question Type} & \multirow{2}*{Setting}& \multicolumn{2}{c|}{FLAN-T5-L} & \multicolumn{2}{c|}{ChatGPT} & \multicolumn{2}{c|}{T5-SFT} & \multicolumn{2}{c}{Ours} \\ 
\cline{4-11}
 \multirow{2}{*}{}&\multirow{2}{*}{} &\multirow{2}{*}{} &EM&F1&EM&F1&EM &F1&EM &F1\\
\midrule
\multirow{4}*{TempReason} &\multirow{2}*{L2: Time-Event} & ReasonQA &  57.3 &66.3&47.5&51.0&82.6&87.1&\textbf{94.1}&\textbf{95.4}\\
\cline{3-11}
\multirow{4}*{} &\multirow{2}*{} & OBQA &  9.4&22.5&8.5&16.1&14.8&35.2&\textbf{19.3}&\textbf{39.1}\\
\cline{2-11}
\multirow{4}*{} &\multirow{2}*{L3: Event-Event} & ReasonQA &  36.3&47.5&49.5&52.3&78.2&83.0&\textbf{93.6}&\textbf{94.4}\\
\cline{3-11}
\multirow{4}*{} &\multirow{2}*{} & OBQA &  8.1&19.2&17.0&25.3&19.7&31.2&\textbf{24.9}&\textbf{35.8}\\
\cline{1-11}

\multirow{2}*{TimeQA} & Easy & OBQA & 35.0 & 44.4 & 43.5 & \textbf{54.0} & 45.1 & 53.7 & \textbf{48.1} & 52.1 \\
\cline{2-11}
\multirow{2}*{} & Hard & OBQA & 27.0 & 35.4 & 33.2 & 43.4 & 36.6 & \textbf{45.5} & \textbf{39.3} & 44.3\\

\bottomrule
\end{tabular} 
}
\vspace{-2mm}
\caption{\label{results}
Comparison of different large language models on varying TSQA datasets.
}
\vspace{-4mm}
\end{table*}
\paragraph{Training}

In the first stage of training , we train T5-base model for 6 epochs with a batch size of 8 on a NVIDIA Tesla V100 GPUs. The model is optimized using AdamW \cite{loshchilov2017decoupled} with a learning rate of $5e^{-6}$. In the second stage of training, we fine-tune the model with our framework for 10 epochs with a batch size of 16. The model is optimized using Proximal Policy Optimization~\cite{schulman2017proximal}. 
In our configuration of the Temporal Information-Aware Embedding, we employ SpaCy, an open-source Natural Language Processing (NLP) library, as time-aware module to locate temporal information within the provided question and context. The window size of the sliding window is set to 10. Furthermore, in Granular Contrastive Reinforcement Learning, the ratio between remote negative answers and proximal negative answers is established at 1:1.

\paragraph{Evaluation Metrics}
We evaluate the model on the test set by exact match (EM) and F1 score, which is the standard evaluation metrics on TempReason and TimeQA. EM is 1 only if prediction and ground truth achieve an exact match and otherwise 0. The value range of F1 is $[0,1]$, and the closer F1 is to 1, the better performance of the model is.

\paragraph{Baselines}

\begin{itemize}
\vspace{-1mm}
    \item FLAN-T5-Large~\cite{wei2021finetuned} represents an advanced instantiation of NLP models based on the extensive T5 framework. 
    We have deployed FLAN-T5-Large for the purpose of evaluating its performance across a variety of TSQA datasets.
\vspace{-1mm}
    \item ChatGPT~\cite{ouyang2022training}, developed by OpenAI, is a sophisticated NLP model predicated on the GPT (Generative Pre-trained Transformer) architecture. 
    Utilizing the official API provided for gpt-3.5-turbo, we have conducted evaluation on various TSQA datasets.
\vspace{-1mm}
    \item T5-SFT~\cite{raffel2020exploring}, an acronym for ``Text-to-Text Transfer Transformer'', is a NLP model conceived by the Google Research team. 
    We subjected the TSQA dataset to supervised fine-tuning on the T5-base model to conduct further evaluation.


\end{itemize}

\subsection{Experimental Results}

Table~\ref{results} contrasts the performance of various LLMs across four TSQA datasets, evaluating them in terms of EM and F1 scores. From the insights drawn from Table~\ref{results}, it is evident that:
1)The framework we propose demonstrates superior performance on the TSQA task, outperforming a range of LLMs, including FLAN-T5-Large, ChatGPT, and T5-SFT models.
2)On these four TSQA datasets, which vary in difficulty levels pertaining to the expression of temporal information and temporal reasoning, our framework shows notable enhancements. Compared to the most competitive baseline results, EM improvements on the L2 and L3 of TempReason are 12.2\% (ReasonQA), 30.4\% (OBQA) and 19.7\% (ReasonQA), 26.4\% (OBQA) respectively, while on the TimeQA dataset's easy and hard versions, improvements of EM are also observed at 6.7\% and 7.4\%, respectively. In the TimeQA dataset, a portion of the answers consists of empty strings. When employing our method, there is a tendency to generate empty strings, which results in a decrease in recall and, subsequently, a reduction in the F1 score. However, by sacrificing a small amount of F1, we can significantly enhance the EM score.

These outcomes validate the exceptional capabilities of our proposed framework in addressing the TSQA task. Regardless of whether the questions contain explicit or implied temporal information, or whether the context is complex or straightforward, our framework consistently demonstrates outstanding performance. For datasets with simpler data (e.g., L2/L3-ReasonQA), our framework aids models in better capturing variations in temporal information within questions and focuses on locating temporal details within brief contexts. For more complex datasets (e.g., L2/L3-OBQA, TimeQA), our framework facilitates the model's ability to focus on and locate temporal information within long contexts that contain multiple unrelated facts and temporal connections. Moreover, it effectively eliminates interference from negative answers of varying granularity, thereby achieving substantial improvements. 
For qualitative results of ``T5-SFT'' and ``Ours'', refer to Appendix~\ref{sec:appendix3}.

\begin{table}[t]

\centering
\resizebox{0.8\columnwidth}{!}{

\begin{tabular}{l|p{2cm}<{\centering}|p{2cm}<{\centering}}
\toprule
\multirow{2}*{Method} & \multicolumn{2}{c}{TemReason-L2}\\ 
\cline{2-3}
 \multirow{2}{*}{} &EM&F1\\
\midrule
T5-SFT &14.8 &35.2 \\
\cline{1-3}
 T5-SFT with TIAE & 18.3 & 37.7\\
\cline{1-3}
 T5-SFT with GCRL &19.2 &38.9 \\
\cline{1-3}
Ours &\textbf{19.3} &\textbf{39.1}\\
\bottomrule
\end{tabular} 
}
\vspace{-2mm}
\caption{\label{ablation1}
Ablation results of model variants with ``OBQA'' setting to explore the contributions of TIAE and GCRL. ``TIAE'' and ``GCRL'' denote ``Temporal Information-Aware Embedding'' and ``Granular Contrastive Reinforcement Learning'', respectively.
}
\vspace{-1mm}
\end{table}

\begin{table}[t]

\centering
\resizebox{\linewidth}{!}{
\begin{tabular}{l|c|p{1cm}<{\centering}|p{1cm}<{\centering}}
\toprule
\multirow{2}*{Method} &\multirow{2}*{Negative Answers} & \multicolumn{2}{c}{TempReason-L2}\\ 
\cline{3-4}
 \multirow{2}{*}{} &\multirow{2}{*}{} &EM&F1\\
\midrule
T5-SFT & Null & 14.8 & 35.2 \\
\cline{1-4}
 T5-SFT with EM-RL & Remote\& Proximal& 16.7 & 37.4 \\
\cline{1-4}
\multirow{3}{*}{T5-SFT with Contrastive-RL} & Remote& 19.0 & 38.7 \\
\cline{2-4}
\multirow{3}{*}{} & Proximal& 18.9 & 38.6\\
\cline{2-4}
\multirow{3}{*}{} & Remote\& Proximal&\textbf{19.2} & \textbf{38.9}\\
\bottomrule
\end{tabular} 
}
\caption{\label{ablation2}
Ablation experiments conducted for the ``Granular Contrastive Reinforcement Learning'' strategy with ``OBQA'' setting, which to explore the impact of various reward functions and the granularity of negative answers on the model's performance.
}
\vspace{-3mm}
\end{table}

\begin{table}[t]
\centering
\resizebox{\linewidth}{!}{
\begin{tabular}{l|c|c|c|c|c|c}
\toprule
\multirow{2}*{Dataset} & \multirow{2}*{Question Type} & \multirow{2}*{Setting} & \multicolumn{2}{c|}{TempT5} & \multicolumn{2}{c}{Ours} \\ 
\cline{4-7}

 \multirow{2}{*}{}&\multirow{2}{*}{} &\multirow{2}{*}{} &EM&F1&EM&F1\\
\midrule
\multirow{4}*{TempReason} &\multirow{2}*{L2: Time-Event} & ReasonQA & 84.8 &88.9 &\textbf{94.1}&\textbf{95.5}\\
\cline{3-7}
\multirow{4}*{} &\multirow{2}*{} & OBQA &15.4 &36.3 &\textbf{19.3}&\textbf{39.1}\\
\cline{2-7}
\multirow{4}*{} &\multirow{2}*{L3: Event-Event} & ReasonQA & 81.1 &86.1 &\textbf{93.6}&\textbf{94.4}\\
\cline{3-7}
\multirow{4}*{} &\multirow{2}*{} & OBQA & 21.1 &32.4 &\textbf{24.9}&\textbf{35.8}\\

\bottomrule
\end{tabular} 
}

\caption{\label{ablation3}
Comparison with TempT5~\cite{tan2023towards}.
}
\vspace{-3mm}
\end{table}

\subsection{Ablation Study}

\paragraph{The Contributions of TIAE and GCRL}
In order to further investigate the contributions of the two methods brought by our framework, we conduct ablation studies by building two more model variants upon ``T5-SFT'':

\begin{itemize}
\vspace{-2mm}
    \item \textbf{T5-SFT with TIAE}, which only applies Temporal Information-Aware Embedding.
\vspace{-2mm}
    \item \textbf{T5-SFT with GCRL}, which only applies Granular Contrastive Reinforcement Learning.
\end{itemize}

The experimental results presented in Table~\ref{ablation1} demonstrate the efficacy of the Temporal Information-Aware Embedding and the Granular Contrastive Reinforcement Learning methods. Specifically, the application of the Temporal Information-Aware Embedding technique alone (T5-SFT with TIAE) resulted in an improvement of 23.6\% (EM) and 7.1\% (F1) on TempReason-L2 dataset. When the model employed solely the Granular Contrastive Reinforcement Learning method (T5-SFT with GCRL), an enhancement of 29.7\% (EM) and 10.5\% (F1) was observed. Moreover, the integration of both strategies yielded a superior performance, culminating in a 30.4\% (EM) and 11.1\% (F1) improvement.

\paragraph{The Novelty of GCRL}

For ``Granular Contrastive Reinforcement Learning'', we explore the impact of the EM score and contrastive triple score as reward functions on the model, as well as the effect of negative answers at varying granularities. We conduct ablation studies by building two additional model variants based upon ``T5-SFT'':

\begin{itemize}
\vspace{-1mm}
    \item \textbf{T5-SFT with EM-RL}, which applies Granular Contrastive Reinforcement Learning, regarded EM score as reward function.
\vspace{-1mm}
    \item \textbf{T5-SFT with Contrastive-RL}, which applies Granular Contrastive Reinforcement Learning, regarded contrastive triple score as reward function.
\end{itemize}

Firstly, in terms of the impact of different reward functions on the model, our observations from Table~\ref{ablation2} indicate that although the model exhibits certain improvements when the EM score is employed as the reward function for reinforcement learning, it still harbors drawbacks, which we have discussed in Section~\ref{sec:3.3}. However, contrastive reinforcement learning mitigates the shortcomings of the EM reward function, achieving a 29.7\% and 10.5\% enhancement of EM and F1.

Secondly, we note the influence of negative answers of varying granularities on the model. We conducted ablation experiments, wherein the negative answers in ``Granular Contrastive Reinforcement Learning'' were set as Remote \& Proximal negative answers, Remote negative answers, and Proximal negative answers, with the quantity of negative answers being consistent across experiments. The results from Table~\ref{ablation2} demonstrate that Remote negative answers and Proximal negative answers contribute to a 28.4\% and 27.7\% improvement of EM in the model, respectively. The combination of both yields a superior performance, culminating in a 29.7\% improvement.

\paragraph{Comparison with TempT5}

Furthermore, we conducted comparative experiments with another framework named TempT5~\cite{tan2023towards} specifically designed for the TempReason dataset, as shown in Table~\ref{ablation3}. Our framework demonstrated a markedly superior performance.




\section{Related Work}

\paragraph{Time Sensitive Question Answering}
Recent developments in Open Book Question Answering (OBQA) have brought about significant advancements in addressing time-sensitive questions. These advancements are particularly evident in contemporary Time Sensitive Question Answering (TSQA) datasets ~\cite{zhang2021situatedqa,chen2021dataset,tan2023towards,yang2024continual}. 

SituatedQA~\cite{zhang2021situatedqa} stands out as a notable contribution to the field. This dataset focuses on open-domain time-sensitive QA, featuring realistic questions reannotated from the NQ dataset ~\cite{kwiatkowski2019natural} to incorporate context dependence and diverse answers across temporal or geographical contexts.

Similarly, TimeQA~\cite{chen2021dataset} presents a dataset comprising 20K questions with an emphasis on time sensitivity. Notably, TimeQA's hard version challenges models to reason over implicit temporal mentions within passages, a feature less emphasized in SituatedQA.

Tan's work~\cite{tan2023towards} in this domain is exemplified by the development of the TempReason benchmark for TSQA. This benchmark offers a comprehensive assessment covering various types of temporal understanding, including time-time (L1) relations, time-event (L2) relations, and event-event (L3) relations. 

These TSQA datasets adopt the OBQA setting, which leverages external context, such as natural language text, to assist Language Models (LMs) in answering questions. This contrasts with Closed Book Question Answering (CBQA), where only the question is provided to the LM without access to external text ~\cite{fevry2020entities,roberts2020much,dhingra2022time,liska2022streamingqa,wei2023menatqa}.

Overall, TSQA tasks highlight the complexity and realism inherent in time-sensitive QA, emphasizing the necessity for LMs to comprehend temporal relationships and effectively ground information temporally to provide accurate responses.

\paragraph{Large Language Models on TSQA}

Some studies have shown that large language models (LLMs) such as those developed by \cite{devlin2018bert}, \cite{raffel2020exploring}, and \cite{liu2019roberta} exhibit strong performance in general question answering tasks \cite{rajpurkar2016squad,kwiatkowski2019natural}. However, with the emergence of Time-Sensitive Question Answering (TSQA) tasks and related datasets, LMs have demonstrated less satisfactory performance. For instance, ~\cite{chen2021dataset} demonstrated that large language models like FiD~\cite{izacard2020leveraging} and BigBird~\cite{zaheer2020big}, functioning as generative and extractive QA models, achieve only 60\% accuracy on TSQA. Furthermore, experimental results from ~\cite{tan2023towards} showed that FLANT5-Large~\cite{wei2021finetuned} and ChatGPT~\cite{ouyang2022training} achieve performance of only around 10\%.

To address the challenges of TSQA, ~\cite{jia2018tequila} proposed TEQUILA, a method specifically tailored for temporal knowledge graph-based QA. TEQUILA utilizes constraint reasoning on temporal intervals to compute answers to question. ~\cite{faghihi2021time} introduced the Time-stamped Language Model as a novel approach to comprehend the progression of events. Their work extends the understanding of event flows across time.
Additionally, Inspired by work on relational KGQA~\cite{huang2019knowledge,saxena2020improving}, ~\cite{shang2022improving} proposed a method of improving time sensitivity for question answering over temporal knowledge graphs~\cite{jia2018tempquestions,jia2021complex,saxena2021question}, which incorporates contrastive learning, time-sensitive temporal KG embedding, and KG-based pruning techniques. ~\cite{mavromatis2022tempoqr} introduced a novel framework by proposing a joint model which integrates temporal knowledge graph embeddings with pre-trained language models. These methods have shown excellent performance on Temporal Knowledge Graphs data. However, as far as we know, there is a lack of effective methods for handling free-text TSQA datasets.
\section{Conclusion}
\vspace{-2mm}
In conclusion, TSQA presents unique challenges for large language models (LLMs), particularly in their attention to and comprehension of temporal information within questions and contexts, as well as in their capacity for temporal reasoning. To bridge these gaps, we have proposed a novel framework that introduces Temporal Information-Aware Embedding and enhances temporal reasoning through Granular Contrastive Reinforcement Learning. This framework notably improves the model's proficiency in discerning relevant temporal details and temporal reasoning.

The comparative evaluation of our framework against existing LLMs across four diverse TSQA datasets has yielded promising results. It not only underscores the enhanced performance of our framework but also highlights the critical need for specialized strategies in addressing the unique demands of TSQA.

\section{Limitation}
Despite these advancements, the performance gap between LLMs and human benchmarks remains an ongoing challenge, suggesting the necessity for continued research and development in this area. Our work lays the groundwork for future explorations into the intricacies of time-sensitive processing and reasoning, aiming to narrow this gap and to elevate the capabilities of LLMs in handling the dynamic and nuanced nature of temporal information.

\section{Ethics Statement}

In conducting our research on Time-Sensitive Question Answering (TSQA), we are acutely aware of the ethical responsibilities we bear, including data privacy and the potential for misinformation.

First and foremost, our work strictly adheres to privacy and data protection standards. The datasets utilized, derived from publicly available information. We ensure that no personal data is used without clear consent and that all information is handled in compliance with relevant data protection laws.
Misinformation is a critical concern in any information processing system. By focusing on temporal accuracy, we aim to reduce the spread of misinformation and improve the reliability of automated responses. However, we acknowledge the inherent limitations of current technologies and the ongoing need for human oversight in verifying the correctness and appropriateness of model outputs. 
Our research team is committed to engaging with these ethical considerations, promoting transparency in our methodologies, and contributing to the responsible development and deployment of TSQA technologies.

\section{Acknowledgements}
This project is partially supported by ARC DP240101349.

\bibliography{custom}

\clearpage
\begin{center}\large\bfseries
Appendix
\end{center}

\appendix

\section{Details of PPO}
\label{sec:appendix1}

Proximal Policy Optimization (PPO) is a widely employed policy gradient method within the domain of reinforcement learning. We utilize the PPO strategy to maximize cumulative rewards. When an action yields negative outcomes, the corresponding reward signal is typically negative. This implies that the action will be reduced in the future. Specifically, the objective of policy is to increase the probability of actions that result in positive rewards and to decrease the probability of actions that lead to negative rewards. Through this mechanism, the policy learns to avoid negative results and gravitates towards positive results. The parameter settings of PPO used in our paper are illustrated in Table~\ref{Parameters1}.

\begin{table}[h]

\centering
\resizebox{0.5\columnwidth}{!}{
\begin{tabular}{l|c}
\toprule
\textbf{Parameters} & \textbf{} \\
\midrule
num\_rollouts & 256 \\ 
chunk\_size& 12  \\
ppo\_epochs& 4   \\
init\_kl\_coef & 0.05  \\
target &6 \\
horizon&10000 \\
gamma &0.99    \\
lam &0.95    \\
cliprange &0.2   \\
vf\_coef &1.0   \\
\bottomrule
\end{tabular} }
\caption{\label{Parameters1} PPO Parameters.}
\end{table}

\section{Examples of Negative Answers with Different Granularity}
\label{sec:appendix2}
In our study, Table~\ref{neg_ans} illustrates several representative examples of remote negative answers and proximal negative answers, which are crucial components in evaluating the effectiveness of our model's temporal reasoning ability. These examples are derived from carefully designed questions and detailed contexts. By incorporating these types of negative answers, the model is challenged to improve its temporal accuracy and enhance its ability to identify correct answers among closely related incorrect options.

\begin{table*}[t]
\begin{tabular*}{\linewidth}{l}
\toprule
\textbf{Question}: Which employer did Mary Warnock, Baroness Warnock work for in 1987?\\
\textbf{Context}: \\From 1949–66, Warnock was a fellow and tutor in philosophy at \textcolor{blue}{St Hugh's College}. \\
...\\
From 1966 to 1972, she was Headmistress at the \textcolor{blue}{Oxford High School} for Girls. \\
...\\
She was Talbot Research Fellow at \textcolor{blue}{Lady Margaret Hall} from 1972 until 1976.\\
...\\
She served as mistress of \textcolor{red}{Girton College}, Cambridge from 1984–1991. ...From 1984 to 1989, Warnock \\chaired a \textcolor{orange}{Home Office Committee} on animal experimentation. ...She was awarded an honorary \\D.Litt. degree by the \textcolor{orange}{University of Bath} in 1987. ...In the 1980s and 1990s, she wrote a column ... and \\gave the \textcolor{orange}{Richard Dimbleby Lecture} on the topic, "Teacher Teach Thyself". \\

\midrule
\textbf{Question}: Who was the head coach of the team Maccabi Haifa F.C in 2013?\\
\textbf{Context}: 
\\
In the 2003–04 season Maccabi Haifa, led by ex-Maccabi footballer \textcolor{blue}{Ronny Levy}, won the championship.\\
...\\
Haifa opened the 2008–09 season with hopes of remaining a top team. At the end of the season, \\coach \textcolor{blue}{Elisha Levy} won his first personal title.\\
...\\
The 2011–12 season was a disappointing one. ...The disappointing season ... and club icon \textcolor{blue}{Reuven Atar} \\as the manager for the next season.\\
...\\
Atar was fired and replaced by \textcolor{red}{Arik Benado}, the youth team coach and former team captain. ...Haifa \\gave fight to \textcolor{orange}{Maccabi Tel Aviv} but finished 2nd. On 25 July 2013, Maccabi Haifa defeated \\\textcolor{orange}{Xazar Lankaran}. ...In 2013, Maccabi Haifa played in the Europa League against \textcolor{orange}{PAOK (Greece)}, \\\textcolor{orange}{AZ Alkmaar (Netherlands)}, and \textcolor{orange}{Shakhter Karagandy (Kazakhstan)}. ...\textcolor{orange}{Sammy Ofer Stadium} was \\completed in November 2013.\\
In 2020–21, Maccabi won the Premier League, their first championship in a decade. Maccabi Haifa \\is currently coached by \textcolor{blue}{Barak Bakhar}.\\

\midrule
\textbf{Question}: Which employer did George Abram Miller work for in 1923?\\
\textbf{Context}: 
\\
in 1890, Miller was Principal of schools in Greeley, Kansas and then professor of mathematics as \\ \textcolor{blue}{Eureka College}.\\
...\\
in 1892. He then joined Frank Nelson Cole at \textcolor{blue}{University of Michigan}.\\
...\\
In 1897 he went to \textcolor{blue}{Cornell University} as an assistant professor, and in 1901 to \textcolor{blue}{Stanford University} \\as associate professor.\\
...\\
In 1906 he went to \textcolor{red}{University of Illinois} where he taught until retirement in 1931. \\...\textcolor{orange}{the Academy of Science of Cracow} awarded a prize and Miller came to prominence in the \\mathematical world abruptly. Miller was president of \textcolor{orange}{the Mathematical Association of America} \\1921–1922 and gave a plenary address at \textcolor{orange}{the International Congress of Mathematicians} in 1924 \\in Toronto.\\
\bottomrule
\end{tabular*}
\caption{\label{neg_ans} Examples of negative answers with different granularity based on question and context. \textcolor{red}{Red} indicates ground truth, \textcolor{blue}{blue} indicates remote negative answers and \textcolor{orange}{orange} indicates proximal negative answers.}
\end{table*}

\section{Qualitative Results}
\label{sec:appendix3}
Table~\ref{Qualitative Results} provides illustrative examples of answers generated by the T5-SFT model compared to those produced by our proposed model, labeled as Ours. This comparison highlights the differences in response quality and accuracy, demonstrating the effectiveness of our approach in handling complex question-answering tasks.

\begin{table*}[t]
\begin{tabular*}{\linewidth}{l}
\toprule
\textbf{Context}: When Jordan left Tynecastle, Snodin returned South to join Barnsley in July 1993, spending \\two seasons in the First Division, before a move to Gainsborough Trinity, where he saw out his \\final playing years with the club, retiring in the summer of 1997.Snodin initially began coaching \\youngsters whilst playing at Gainsborough Trinity as he ran the \"Glynn Snodin Soccer Academy\" at \\Gainsborough Leisure Centre on Saturday mornings between 1995 and 1997.\\
\textbf{Question}: Which team did Glynn Snodin play for in Jul, 1996?\\
\textbf{T5-SFT}: ``Leeds United F.C.'' \\
\textbf{Ours}: ``Gainsborough Trinity F.C.''\\
\textbf{Ground Truth}: ``Gainsborough Trinity F.C.''\\
\midrule
\textbf{Context}: Since 2017, he is director of Spotify's Creator Research Technology Lab in Paris, where \\he develops tools for assisting music creation.The Music team at Sony Computer Science Laboratory \\Paris was founded in 1997 by Pachet, where he developed the vision that metadata can greatly enhance \\the musical experience, from listening to performance.The Flow Composer is his second achievement, \\a system to compose lead sheets in the style of arbitrary composers.\\
\textbf{Question}: Which employer did Fran Pachet work for in Sep, 2019?\\
\textbf{T5-SFT}: ``Sony Computer Science Laboratories''\\
\textbf{Ours}: ``Spotify''\\
\textbf{Ground Truth}: ``Spotify''\\
\midrule
\textbf{Context}: In 1917 he journeyed through the Middle East with Archduke Hubert Salvator of Austria; \\there are suggestions that the mission had a political motive involving Arab Revolt against the \\Ottoman government.After the war he became a professor at Charles University in Prague (1920), \\despite opposing voices resenting his close ties with the House of Habsburg. He helped to establish \\the Oriental Institute of the Academy of Sciences in Prague.In cooperation with the American \\industrialist Charles Richard Crane he published his works in English (1922-2013). In addition to \\scientific work and popular travel books he published 21 novels for young readers. Musil worked \\for Charles University until 1938, but was active until the very end of his life. \\
\textbf{Question}: Which employer did Alois Musil work for in Oct, 1922?\\
\textbf{T5-SFT}: ``Oriental Institute of the Academy of Sciences''\\
\textbf{Ours}: ``Charles University''\\
\textbf{Ground Truth}: ``Charles University''\\
\midrule
\textbf{Context}: Stanislav GrossStanislav Gross (; 30 October 1969-2013 16 April 2015) was a Czech lawyer \\and politician who served as Prime Minister of the Czech Republic and Leader of the Czech Social \\Democratic Party from 2004 until 2005 when he resigned as a result of his financial irregularities. \\He previously served as Minister of the Interior in cabinets of Miloš Zeman and Vladimír Špidla \\from 2000 to 2004. \\
\textbf{Question}: Which position did Stanislav Gross hold in Oct, 2001?\\
\textbf{T5-SFT}: ``Member of the Chamber of Deputies of the Czech Republic''\\
\textbf{Ours}: ``Minister of the Interior of the Czech Republic''\\
\textbf{Ground Truth}: ``Minister of the Interior of the Czech Republic'' \\

\bottomrule
\end{tabular*}
\caption{\label{Qualitative Results} Examples of answers generated by ``T5-SFT'' and ``Ours''.}
\end{table*}
\end{document}